# Transferring Adaptive Theory of Mind to social robots: insights from developmental psychology to robotics

Francesca Bianco[1*] and Dimitri Ognibene[1]

[1] University of Essex, Colchester CO4 3SQ, UK
*fb18599@essex.ac.uk

**Abstract.** Despite the recent advancement in the social robotic field, important limitations restrain its progress and delay the application of robots in everyday scenarios. In the present paper, we propose to develop computational models inspired by our knowledge of human infants' social adaptive abilities. We believe this may provide solutions at an architectural level to overcome the limits of current systems. Specifically, we present the functional advantages that adaptive Theory of Mind (ToM) systems would support in robotics (i.e., mentalizing for belief understanding, proactivity and preparation, active perception and learning) and contextualize them in practical applications. We review current computational models mainly based on the simulation and teleological theories, and robotic implementations to identify the limitations of ToM functions in current robotic architectures and suggest a possible future developmental pathway. Finally, we propose future studies to create innovative computational models integrating the properties of the simulation and teleological approaches for an improved adaptive ToM ability in robots with the aim of enhancing human-robot interactions and permitting the application of robots in unexplored environments, such as disasters and construction sites. To achieve this goal, we suggest directing future research towards the modern cross-talk between the fields of robotics and developmental psychology.

**Keywords:** Mental States, Computational Modelling, Theory of mind

## 1 Introduction

The robotic field has greatly advanced in the last decades and the complexity and sophistication of robots have highly improved. However, the application of robotics in everyday scenarios still faces some challenges. For example, humans' positive attitude towards non-human agents has increased with the humanoid aspect of robots, however, this drops when a mismatch of agent features is noticed [1] due to the robots' still limited (non-human-like) social capabilities (uncanny valley effect). Currently, robots' architectures generally rely on association and simulation principles to learn about the social world, which enable them to recognize and predict actions from observing other agents performing such actions [2, 3]. However, robots' recognition of complex mental states is still limited, as is their understanding of the humans they interact with. For example, although robots can recognize the action of digging through rubble, they do not necessarily understand the deeper mental state of the observed agent, such as the desire to search for survivors.

In robotics, the perception of social contexts mostly relies on the passive view of stimuli (with few exceptions [4]) and highly prewired knowledge. This means that missing elements and features are not actively searched by robots and that their adaptability to new contexts/tasks/agents' behaviors is limited. Similarly, multiple-agent interactions may be too computationally and sensorially expensive given the inability to exploit context-specific information to optimize the computations underlying robots' social interactions. Therefore, one of the main challenges in robotics is to create robots that can act as social (human-like) agents to increase robots' acceptance as social companions and ameliorate human-robot interactions, also in challenging situations [5].

Substantial advancements in robotics resulted from the introduction of Deep Neural Networks (DNNs). However, online action perception, online learning and generalization to new contexts with a sufficient level of spatial and temporal detail to support human-robot interaction is still difficult [6]. One of the main issues related to the application of DNNs for social interaction is their reliance on suitable datasets. Current DNNs algorithms require huge datasets which are mostly recorded by humans, who however select points of view which may completely differ from those adopted by robots. While the performance of these bottom-up recognition methods is continuously improved through new architectures [7] and datasets [8], the high dependency of human activities on multiple contextual factors and actors' mental states suggests that the size of the datasets necessary to achieve a high enough precision for predictive physical interaction would be prohibitive. Furthermore, this approach requires substantial training time and reconfiguration or retraining when new tasks are added.

An initial solution to these issues was provided by simulation-based methods for the extension of datasets through the generation of virtual training data [9] and digital manipulation of the training data [10]. However, these methods are particularly difficult to scale to social interaction tasks because both the virtual training data and the robot's response are strongly affected by contexts of high dimensionality (e.g., other agent's previous actions, current posture, relative position of other objects). The DNNs approach is also limited given that learning is based on short frames and does not allow for an interpretation of the situation or the agent's mental state. Therefore, this only enables the understanding of short and stereotyped interactions. This is in contrast to humans' ability to comprehend the world by adding sequences together, putting them into context and relating them to other similar events or their own experiences, feelings, mental states. Ultimately, the lack of data in uncertain, previously unexperienced environments (e.g., natural disasters) prevents the utilization of standard machine learning methods to provide robots with the correct behaviors.

A way to deal with the strong limits of DNNs and endow robots with social skills would be to integrate them in a principled manner with a model-based reasoning system [11]. Indeed, we propose to develop future computational models inspired by our knowledge of human infants' mentalizing abilities, as they may provide solutions at an architectural level to ameliorate the limits of existing robotic systems. In this paper, we present the functional advantages that adaptive mentalizing systems would support in robotics and contextualize them in practical applications. Specifically, we review current models and robotic implementations to identify the current limitations and suggest a possible future developmental pathway.

## 2  Adaptive Theory of Mind for Robots

Theory of Mind (ToM), or mentalizing, refers to the cognitive capacity to attribute and represent others' mental states [12-14]. Although a definite age at which mentalizing develops in humans is yet to be determined, possessing a ToM from an early age [15] is an evident demonstration of its importance for human social navigation. In fact, "externally observable actions are just observable consequences of unobservable, internal causal structures" [16], i.e., mental states. In other words, if a person is running down a mountain, his immediate goal is to displace himself from a location to another, whereas his underlying intention (e.g., to run away from an avalanche) is not clear until the context is analyzed. To a certain extent, humans are able to predict and understand others' actions by observing their behavior, however, having the capability to infer the underlying reasons provides an invaluable advantage during social interactions. Given that mental states are characterized differently in the literature (see Table 1 for further details), a clear definition of ToM is yet to be identified. However, compared to the association, simulation and teleological paradigms underlying social perception, the highlight of ToM relies on the quantities it extracts (i.e., mental states) rather than the computational processes that realize the inferences. A better understanding of how these methodologies can support ToM is provided in section 3.

**Table 1.** Different characterization of mental states in the literature.

| Types of mental states |
| --- |
| Intentions, Desires and Beliefs [21] |
| Desires, Beliefs and Percepts [23] |
| Desires, Values, Beliefs and Expectations [24] |
| Perceptions, Bodily feelings, Emotional states and Propositional attitudes (Beliefs, Desires, Hopes and Intentions) [17] |
| Long-term dispositions, Short-term emotional states, Desires and their associated goal-directed Intentions and Beliefs about the world [18] |

### 2.1  Functional advantages for robotics

Equipping robots with a ToM would allow them to also access human's hidden mental states, such as intentions, desires and beliefs, and to reason about and react to them much like humans do. More specifically, it would facilitate the adaptive attribution of mental states to agents, meaning for example that beliefs would be acquired and intentions would be inferred by the robot itself. By equipping robots with an adaptive ToM, their application to situations in which specific data are not currently available, due to high variability or uncertainty of environments/agents, will be possible (e.g., searching and rescuing during disasters or helping in construction sites). In fact, internal beliefs/mental states are generally shared by humans and provide important context for higher detail and perceptually-demanding behavior understanding. Finally, providing robots with mentalizing abilities would enable them to "express their internal states through social interaction" [17], which will answer to the issue expressed by the uncanny valley effect. The functional advantages of an adaptive ToM for robotics (see paper [18] for further details) will now be described (Table 2).

Table 2. Functional advantages of equipping robots with a Theory of Mind.

| Functional advantages of ToM for Robotics | Current systems | ToM contribution |
| --- | --- | --- |
| *Mentalizing for coordinating and managing false beliefs* | - Spatial perspective taking<br>- Robot has access to states and actions of all agents (≠ real-life) | - Mental perspective taking |
| *Mentalizing for proactivity and preparation* | - Reliance on bottom-up inputs | - Top-down control: proactivity<br>- Improved preparation for interaction with other agents |
| *Mentalizing for active perception* | - Passive exploration/understanding of environments/agents | - Active search of cues: better explain current situation and enable more precise predictions |
| *Mentalizing for learning* | - Deep neural network limits<br>- Same learning process for passive object dynamics and human behavior | - Time- and cost-efficient learning<br>- Different learning processes: object dynamics VS human behavior<br>- Improved multi-agent interactions |

**Mentalizing for coordinating and managing false beliefs.** Until now, ToM has mainly been implemented in robotics to allow for the understanding and ability to track beliefs in humans [19, 20]. In fact, determining whether a human partner holds true beliefs about a situation is an essential requirement for successful human-robot interactions [20], especially during collaborative tasks. Previous studies introduced in robots the ability to assume the spatial perspective of the agent they were interacting with [19, 21], which is a fundamental characteristic of human mentalizing. By enabling robots to put themselves in the agents' shoes and infer their sensorial access, they showed a better recognition of mental states and increased performances in belief recognition tasks. Interestingly, false-belief tasks (standard experimental paradigm to test ToM in humans [22]) were implemented in robots [20] and were passed with the aid of spatial perspective taking. Future studies should create adaptive ToM architectures which aim at also equipping robots with mental perspective taking. This would mean systems in which robots are able to autonomously attribute a wider set of mental states, reason about them and appropriately react to them. Another approach was recently presented by Rabinowitz et al. [8], who proposed a NN able to predict the behavior of multiple agents in a false-belief situation given their past and current trajectories. However, as the authors mention in their paper, they assume the observer to have access to states and actions of all agents, which is not always the case for an embodied agent. Therefore, fully understanding and reasoning about agents' beliefs still represents a challenge for robotics.

**Mentalizing for proactivity and preparation.** Characterizing other agents through their beliefs, desires and intentions may allow the anticipation of their behavior before they perform any concrete action. Proactivity implies a lower reliance on bottom-up inputs in favor of additional top-down control influencing the response of the robot to the situation. This is important for the successful application of robots in everyday social settings and collaborative tasks, as social contexts are highly dynamic and robots are required to prepare and act prior to an event rather than just to react to it. An interesting study by Milliez et al. [20] provided an example of a proactive robot which was able to reason about the beliefs of a human partner and to communicate important information that the human had missed for the successful completion of a collaborative task. Although the architecture of this robot was based on the ToM principle of perspective taking to interact with an agent, the robot was readily provided with several hard-coded position hypotheses to make when an object was not visible (contrary to the automaticity seen in human behavior). Ultimately, equipping robots with an adaptive ToM would allow more efficient and fluid human-robot interactions (e.g., by independently positioning themselves in a position easy to spot).

**Mentalizing for (active) perception.** Associating intentions and mental states to agents' behavior may encourage observers to search for cues that better explain the current situation and enable more precise predictions [23]. Active perception may be necessary to eliminate the passive nature of robots' exploration and understanding of environments/agents. For example, in Görür et al. [19] the robot was fed with 100 different observation sequences from which states were estimated with a hidden Markov model. This suggests that the important information/features of the scene were readily provided to the robot, which was not left to explore and act in the environment to increase the information content derived from its sensorial data. In contrast with the ecological behavior seen in humans, this limits the quality of human-robot interactions.

**Mentalizing for learning.** An adaptive ToM for robots would also tackle many of the challenges identified in the robotics field by Lake et al. [24]. Integrating ToM development principles in the blueprint of an adapting neural architecture for social interaction may result in a more time- and cost-efficient learning process compared to DNNs [6]. Furthermore, it would decrease the need to select and feed appropriate content to robots through expensive datasets, also reducing human errors involved in their preparation and permitting increased accuracy. In addition, mentalizing for learning would imply a different way of learning about the world. More specifically, most DNN-based action recognition systems do not currently distinguish the learning of passive objects dynamics (e.g., the movement of clouds) from that of agents' behavior (e.g., opening several boxes to find lost glasses while searching) and do not take into consideration the intentionality that marks humans' behaviors. In contrast, we propose that equipping robots with mental states understanding and contextualization would provide a means to distinguish between passive object dynamics and agents behaviors. Finally, predicting interactions and perceiving the mental states of multiple agents simultaneously may be particularly demanding both for the sensory and the general cognitive load of the agent.

However, humans swiftly deal with these conditions by adaptively allocating their attention to the most significant users based on contextualized knowledge (e.g., while playing football, only the intentions of the player with the most relevant location and role will be considered) [25]. Therefore, providing robots with similar capabilities would alleviate the current issues associated with multi-agent interactions.

## 3 Computational modelling of ToM

In the literature, several theories describing human ToM exist [12, 26], however, developing a computational model inspired by the brain processes underlying human mentalizing is particularly challenging. Here, we attempt to better delineate the possible brain processes responsible for the development of the mentalizing ability in humans with the aim of transferring such features to social robots and stimulating future research in the direction proposed. Specifically, we will describe the limitations and advantages of two principal accounts which have been contrasted and implicated in the development of a ToM in humans, i.e., the teleological and simulation theories.

### 3.1 Teleological theory for ToM

The teleological theory is one of the principal accounts utilized to describe the intention recognition ability based on observable actions in both adults and infants [27]. Specifically, infants were suggested to attribute a causal intention to an agent according to the rationality principle [28]. However, whether this teleological account is a suitable candidate mechanism underlying ToM remains questioned. Firstly, the concepts represented in the mentalistic account can be considered more complex compared to those of the teleological account. In fact, although the teleological account is able to process actions to derive the goal of an agent in various situations, it is unlikely that the rationality principle may provide access to the unobservable, abstract mental states [29, 30]. Indeed, a recent review noted that there are kinds of mindreading contexts that have nothing to do with rationality or efficiency [12]. Similarly, rationality, thus the teleological account, is not very effective when trying to infer mental states which are subjective, as efficiency may not be the prerogative of the agent observed. In addition, while the teleological account suggests that infants should not be able to distinguish their representation of a scene from that of an agent (thus, reality should be as construed by the infants), the mentalistic account presupposes the attribution of a perspective to another agent, which may be similar or differ from their own [31].

Nonetheless, Gergely and Csibra [28] proposed a continuum between the teleological constructs (i.e., action, goal-state and situational constraints) and the mentalistic ones, with the latter presupposing the same computations and constructs of the former but representing more sophisticated, abstract constructs (i.e., intentions, desires and beliefs). Furthermore, Baker et al. [32] developed a Bayesian computational model for ToM based on the teleological principle that was tested on both adults and infants [33], who were suggested to follow this model to infer the mental states behind an agent's behavior using priors. However, this model is computationally demanding and could not be directly used to support online interactions.

### 3.2 Simulation theory for ToM

The simulation theory is the other principal account utilized to describe the intention recognition ability based on observable actions in both adults and infants [34, 35]. Specifically, activation of infants' motor system was shown both during the observation of a grasping action and prior to the visual input once the action could be predicted [35]. However, similarly to the teleological account, whether the simulation account is a suitable candidate mechanism underlying ToM remains questioned. The simulation theory "proposes that we can understand the mental states of others on the basis of our own mental states" [12, 13, 34, 36]. Therefore, in contrast to the teleological account, the simulation theory permits the representation of the same abstract mental states, given that we experience our own mental states. However, having the same desire as another person does not necessarily permit the inference of their intentions. Hence the simulation theory can be considered only a first step for mentalizing [13]. Furthermore, the simulation account can account for the subject-specific nature of the mental states only when the observer and the subject observed are very similar [13]. This has also consequences on the metarepresentational ability of mentalizing. In fact, while similarity is essential to permit the transfer of mental perspectives, it may also be a disadvantage and lead to the quarantine failure [12]. That is the failure to both exclude own mental states (which are lacking in the agent observed) and include those possessed by the target (as lacking in the observer). Against the simulation theory as a base for mentalizing is also some evidence of its inability to support action understanding in novel situations [36], which does not support the context-specific nature of mental states.

Nonetheless, Keysers and Gazzola [37] proposed a model integrating the simulation and mentalistic accounts based on neural evidence. More specifically, the authors suggested that the brain areas associated with both accounts reflect simulation, even though at different levels, rather than radically different processes.

### 3.3 Integrating the models for a better ToM

Although the teleological and simulation models in some respects rely on different representations and computations, they may be important in different situations or when dealing with specific mental states. For example, while the teleological model might be useful to predict mental states early in development given its more innate nature (due to the central rationality principle) compared to the simulation model, the latter may become valuable when humans start learning from experience and relating to other people. Similarly, while the simulation approach may be more suitable to infer mental states triggered by bottom-up stimuli, the teleological model may be important when an increasing top-down control is necessitated. In turn, the top-down control enabled by the mentalistic and teleological models may enable different preparation strategies for interaction, such as adopting a convenient posture (e.g., looking for a possible target of a predator before it approaches it [4]).

In this paragraph we suggest an innovative view which is not usually taken in robotics, that is the integration of the simulation and teleological models for ToM as a means to improve robots' social skills. We thus propose a complementary view of the models described up to date, rather than a contrasting one. In fact, although mirroring does not necessarily imply inference and prediction of the final intentions and beliefs of an agent, it may help with the action sequences necessary to reach that goal state (i.e., the trajectory to reach the final state). This may favor the teleological reasoning which may provide further information to infer and predict the mental states of the agent observed. The same might be true also in the opposite direction. While the teleological model might provide information on possible trajectories of observed actions to infer the agent's mental states, simulation may allow the correct inference of intentions, desires or beliefs by choosing between such options through internal simulation.

## 4 Questions and future directions

A great debate currently exists on whether the abstract mental states can be accessed through learning or if they can be innately understood. Based on what mentioned above, it would be reasonable to assume that they can be inferred after learning about an agent through repeated observations, in different contexts, with the aid of language and communication [13, 14, 38]. However, the assumption that this capacity may be innate (i.e., derived from the supposition that conspecifics share general mental states) or driven by innate stimulus cues (e.g., direction of gaze or movement) also seems valid [28, 39]. It is possible that a combination of the two inference mechanisms occurs in humans [40].

Another debate concerns whether such mental states can be inferred directly from automatic, bottom-up effects, such as the automatic tendency to share another person's experiences, or whether mostly top-down control is involved [30]. Again, these recognition mechanisms may act in concert to achieve optimal mental states understanding.

Shedding light on these processes for ToM would also mean assessing which current models are better describing the mentalizing ability and whether they cooperate or compete with each other. Nevertheless, assigning an innate component to and a top-down control over the mental states inference process would support the teleological account as a precursor of ToM. In contrast, if a learning component and a bottom-up control are assumed, the simulation account could be identified as the precursor of ToM. Finally, if all these properties are present during ToM at different instances or when attributing different mental states, both the teleological and simulation models might be precursors of the mentalizing ability. They may however be important for specific parts of ToM.

We would like to urge future studies to focus on the modern cross-talk between developmental studies and robotics to answer these questions. In fact, on the one hand, developing architectures for robots inspired by developmental mechanisms resulted in more sophisticated robots with increasingly complex abilities and behavior [41]. On the other hand, robots have been useful in the modelling of human developmental processes within an embodied agent and the prediction of developmental phenomena which were successively validated by infants studies [2]. Therefore, developing robot architectures based on ToM can result in increasingly complex adaptive social robots as well as in a new tool for investigating models from developmental psychology and provide insights into human capabilities which are yet to be fully understood, including the mentalizing ability itself.

# References

1. Abubshait, A., Wiese, E.: You Look Human, But Act Like a Machine: Agent Appearance and Behavior Modulate Different Aspects of Human-Robot Interaction. Frontiers in psychology 8, 1393 (2017)
2. Cangelosi, A., Schlesinger, M.: From Babies to Robots: The Contribution of Developmental Robotics to Developmental Psychology. Child Development Perspectives 12, 183-188 (2018)
3. Demiris, Y., Dearden, A.: From motor babbling to hierarchical learning by imitation: a robot developmental pathway. Proceedings of the 5th International Workshop on Epigenetic Robotics, pp. 31-37 (2005)
4. Ognibene, D., Demiris, Y.: Towards active event recognition. Proceedings of IJCAI AAAI, pp. 2495-2501 (2013)
5. Wiese, E., Metta, G., Wykowska, A.: Robots as Intentional Agents: Using Neuroscientific Methods to Make Robots Appear More Social. Frontiers in psychology 8, 1663 (2017)
6. Pierson, H., Gashler, M.: Deep learning in robotics: a review of recent research. Advanced Robotics 31, 821-835 (2017)
7. Singh, G., Saha, S., Sapienza, M., Torr, P., Cuzzolin, F.: Online real time multiple spatio-temporal action localisation and prediction on a single platform. arXiv preprint arXiv:1611.08563 (2017)
8. Rabinowitz, N.C., Perbet, F., Song, H.F., Zhang, C., Eslami, S.M.A., Botvinick, M.: Machine Theory of Mind. arXiv preprint arXiv:1802.07740 (2018)
9. Mariolis, I., Peleka, G., Kargakos, A., Malassiotis, S.: Pose and category recognition of highly deformable objects using deep learning. Advanced Robotics (ICAR) International Conference, pp. 655-662 (2015)
10. Polydoros, A.S., Nalpantidis, L., Kruger, V.: Real-time deep learning of robotic manipulator inverse dynamics. Intelligent Robots and Systems (IROS) IEEE/RSJ International Conference, pp. 3442-3448 (2015)
11. Silver, D., Schrittwieser, J., Simonyan, K., Antonoglou, I., Huang, A., Guez, A., et al.: Mastering the game of Go without human knowledge. Nature 550, 354-359 (2017)
12. Goldman, A.I.: Theory of mind. in The Oxford Handbook of Philosophy of Cognitive Science, (Oxford: Oxford University Press) (2012)
13. Frith, C.D., Frith, U.: The neural basis of mentalizing. Neuron 50, 531-534 (2006)
14. Devaine, M., Hollard, G., Daunizeau, J.: The social Bayesian brain: does mentalizing make a difference when we learn? PLoS computational biology 10:e1003992 (2014)
15. Yott, J., Poulin-Dubois, D.: Are Infants' Theory-of-Mind Abilities Well Integrated? Implicit Understanding of Intentions, Desires, and Beliefs. Journal of Cognition and Development 17, 683-698 (2016)
16. Kosakowski, H.L., Saxe, R.: "Affective theory of mind" and the function of the ventral medial prefrontal cortex. Cognitive and Behavioral Neurology 31, 36-50 (2018)
17. Scassellati, B.: Theory of mind for a humanoid robot. Autonomous Robots 12, 13-24 (2002)
18. Bianco, F., Ognibene, D.: Functional Advantages of an adaptive Theory of Mind for robotics: a review of current architectures. The 11th Computer Science and Electronic Engineering Conference. IEEE Xplore, University of Essex (2019)
19. Görür, O.C., Rosman, B., Hoffman, G., Albayrak, A.: Toward Integrating Theory of Mind into Adaptive Decision-Making of Social Robots to Understand Human Intention. In: Workshop on Intentions in HRI at ACM/IEEE International Conference on Human-Robot Interaction (2017)
20. Milliez, G., Warnier, M., Clodic, A., Alami, R.: A framework for endowing an interactive robot with reasoning capabilities about perspective-taking and belief management. The 23rd

IEEE International Symposium on Robot and Human Interactive Communication, pp. 1103-1109 (2014)
21. Devin, S., Alami, R.: An Implemented Theory of Mind to Improve Human-Robot Shared Plans Execution. The Eleventh ACM/IEEE International Conference on Human Robot Interation, pp. 319-326 (2016)
22. Grosse, W.C., Friederici, A.D., Singer, T., Steinbeis, N.: Implicit and explicit false belief development in preschool children. Developmental Science 20:e12445 (2017)
23. Ognibene, D., Chinellato, E., Sarabia, M., Demiris, Y.: Contextual action recognition and target localization with an active allocation of attention on a humanoid robot. Bioinspiration & biomimetics 8:035002 (2013)
24. Lake, B.M., Ullman, T.D., Tenenbaum, J.B., Gershman, S.J.: Building machines that learn and think like people. Behavioral and Brain Sciences 40, 1–101 (2016)
25. Lee, K., Ognibene, D., Chang, H.J., Kim, T.-K., Demiris, Y.: STARE: Spatio-Temporal Attention Relocation for Multiple Structured Activities Detection. IEEE Transactions on Image Processing 24, 5916-5927 (2015)
26. Schaafsma, S.M., Pfaff, D.W., Spunt, R.P., Adolphs, R.: Deconstructing and reconstructing theory of mind. Trends in Cognitive Sciences 19, 65–72 (2015)
27. Southgate, V., Johnson, M.H., Csibra, G.: Infants attribute goals even to biomechanically impossible actions. Cognition 107, 1059-1069 (2008)
28. Gergely, G., Csibra, G.: Teleological reasoning in infancy: the naive theory of rational action. Trends in Cognitive Sciences 7, 287-292 (2003)
29. Koster-Hale, J., Richardson, H., Velez, N., Asaba, M., Young, L., Saxe, R.: Mentalizing regions represent distributed, continuous, and abstract dimensions of others' beliefs. NeuroImage 161, 9-18 (2017)
30. Frith, C.D., Frith, U.: How we predict what other people are going to do. Brain research 1079, 36-46 (2006)
31. Luo, Y., Baillargeon, R.: Toward a Mentalistic Account of Early Psychological Reasoning. Current directions in psychological science 19, 301-307 (2010)
32. Baker, C.L., Jara-Ettinger, J., Saxe, R., Tenenbaum, J.B.: Rational quantitative attribution of beliefs, desires and percepts in human mentalizing. Nature Human Behaviour 1:0064 (2017)
33. Hamlin, J.K., Ullman, T., Tenenbaum, J., Goodman, N., Baker, C.: The mentalistic basis of core social cognition: experiments in preverbal infants and a computational model. Dev Sci 16, 209-226 (2013)
34. Gallese, V., Goldman, A.: Mirror neurons and the simulation theory of mind-reading. Trends in Cognitive Sciences 2, 493-501 (1998)
35. Southgate, V., Johnson, M.H., Osborne, T., Csibra, G.: Predictive motor activation during action observation in human infants. Biology letters 5, 769-772 (2009)
36. Brass, M., Schmitt, R.M., Spengler, S., Gergely, G.: Investigating action understanding: inferential processes versus action simulation. Current biology 17, 2117-2121 (2007)
37. Keysers, C., Gazzola, V.: Integrating simulation and theory of mind: From self to social cognition. Trends in Cognitive Sciences 11, 194-196 (2007)
38. Frith, C.D., Frith, U.: Social cognition in humans. Current biology 17, 724-732 (2007)
39. Kovacs, A.M., Teglas, E., Endress, A.D.: The social sense: susceptibility to others' beliefs in human infants and adults. Science 330, 1830-1834 (2010)
40. Baron-Cohen, S.: Mindreading: Evidence for both innate and acquired factors. Journal of Anthropological Psychology 17, 26-27 (2006)
41. Bhat, A.A., Mohan, V., Sandini, G., Morasso, P.: Humanoid infers Archimedes' principle: understanding physical relations and object affordances through cumulative learning experiences. Journal of the Royal Society, Interface 13 (2016)